\documentclass[journal]{IEEEtran}
\ifCLASSINFOpdf
\else
\fi
\usepackage{amsmath}
\usepackage{graphicx}
\usepackage{booktabs}
\usepackage{float}
\usepackage{caption}
\usepackage{verbatim}
\usepackage{cite}
\usepackage{subcaption}
\usepackage{makecell}
\DeclareMathOperator{\cas}{cas}

\DeclareMathOperator{\swish}{swish}

\DeclareMathOperator{\ReLU}{ReLU}

\DeclareMathOperator{\GeLU}{GeLU}

\begin{document}


\title{Stochastic Multivariate Universal-Radix Finite-State Machine: a Theoretically and Practically Elegant Nonlinear Function Approximator}

\author{{Xincheng Feng,}
        {Guodong Shen,}
        {Jianhao Hu,}
        {Meng Li,}
        {Ngai Wong.}}

\maketitle
\thanks{X. Feng is with the Department of Electrical and Electronic Engineering, the University of Hong Kong. Email: xinchengfeng@yeah.net}

\thanks{G. Shen is with the National Key Laboratory of Wireless Communications, University of Electronic Science and Technology of China. Email: shengd@std.uestc.edu.cn}

\thanks{J. Hu is with the National Key Laboratory of Wireless Communications, University of Electronic Science and Technology of China. Email: jhhu@uestc.edu.cn}

\thanks{M. Li is with the Institute for Artificial Intelligence, Peking University. Email: meng.li@pku.edu.cn}

\thanks{N. Wong is with the Department of Electrical and Electronic Engineering, the University of Hong Kong. Email: nwong@eee.hku.hk}

\begin{abstract}
Nonlinearities are crucial for capturing complex input-output relationships especially in deep neural networks. However, nonlinear functions often incur various hardware and compute overheads. Meanwhile, stochastic computing (SC) has emerged as a promising approach to tackle this challenge by trading output precision for hardware simplicity. To this end, this paper proposes a first-of-its-kind stochastic multivariate universal-radix finite-state machine (SMURF) that harnesses SC for hardware-simplistic multivariate nonlinear function generation at high accuracy. We present the finite-state machine (FSM) architecture for SMURF, as well as analytical derivations of sampling gate coefficients for accurately approximating generic nonlinear functions. Experiments demonstrate the superiority of SMURF, requiring only 16.07\% area and 14.45\% power consumption of Taylor-series approximation, and merely 2.22\% area of look-up table (LUT) schemes.
\end{abstract}


\IEEEpeerreviewmaketitle

\section{Introduction}
Nonlinear functions play a crucial role in artificial intelligence (AI), enabling complex computations and sophisticated modeling. They enhance neural networks by introducing nonlinearity through activation functions such as rectified linear unit ($\ReLU$) and Gaussian error linear unit ($\GeLU$), which allow AI models to capture intricate relationships and improve performance in tasks like image recognition~\cite{1,2,3}, natural language processing~\cite{4,5}, and speech recognition~\cite{6}. Additionally, nonlinear functions are essential in dimensionality reduction techniques like kernel methods and manifold learning~\cite{7,8,9}, uncovering nonlinear patterns in high-dimensional data. Optimization algorithms like gradient descent also rely on nonlinear functions to locate optimal solutions~\cite{10}.

However, incorporating nonlinear functions, such as exponentiation ($\exp$) and logarithm ($\log$), increases hardware complexity and computational overheads. The nonlinear activation at each neuron induces compute and memory consumption, making it crucial to have simple yet accurate nonlinear function generators especially for lightweight edge AI.

Several representative schemes have been proposed for computing univariate nonlinear functions, including the coordinate rotation digital computer (CORDIC) algorithm~\cite{11}, Bernstein-polynomials-based approximation~\cite{12}, Taylor-series approximation~\cite{13}, look-up tables~(LUTs) and stochastic computing (SC)~\cite{14}. In particular, SC has demonstrated high efficiency in nonlinear computations across various applications, such as image and signal processing~\cite{15,16}. Initial investigations on SC implementation were inspired by Gaines' logical computation of random bitstream~\cite{17}. Brown and Card~\cite{14} later proposed an SC-based scheme using finite-state machines (FSMs) to implement specific functions, but it was limited to $\exp$ and $\tanh$ functions. Qian et al. introduced a general method converting nonlinear functions into Bernstein polynomials using random bitstreams~\cite{12}. However, FSM generalization for approximating nonlinear functions remains limited. A method based on multi-driving and multi-dimension FSMs (MM-FSMs) was proposed in 2022~\cite{18}. This approach employs an FSM structure consisting of multiple rows and columns instead of a chain structure, enabling the implementation of primarily univariate activation functions used in AI, digital signal processing, and communication systems on a common hardware platform with low complexity and energy consumption.

Nevertheless, all these methods only handle univariate nonlinear functions, and a gap exists for the design and extension to bivariate functions and beyond. To this end, we propose a first-of-its-kind FSM dubbed \textbf{S}tochastic \textbf{M}ultivariate \textbf{U}niversal-\textbf{R}adix \textbf{F}inite-state machine (SMURF) for general multivariate nonlinear function approximation. Specifically, SMURF represents a general hardware architecture comprising simple circuit components like flip-flops, comparators and multiplexers (MUXs), and offers configurable parameters for generating any multivariate nonlinear functions. The key contributions of this paper are:
\begin{itemize}
\item We propose the \emph{first-ever} SC-based FSM, dubbed SMURF, for multivariate nonlinear function computing.
\item We present the derivation of steady-state probabilities and analytical means of setting coefficient parameters in SMURF for generic nonlinear functions.
\item We conduct software simulation and hardware benchmarking of SMURF. Experiments show that while the proposed method achieves the same output accuracy as the conventional Taylor-series approximation, the power consumption is only 14.4\%, and the area consumption is only 20.9\%. Furthermore, the area consumption of SMURF is only 2.90\% of LUT-based schemes.
\end{itemize}

The rest of this paper is organized as follows. Section II provides the necessary preliminaries of SC and random sampling gates. Section III introduces the SMURF scheme and presents the method to synthesize the parameters. Section IV shows the performance of SMURF in various nonlinear functions in terms of computing accuracies, demonstrates the accuracy performance of an SC-based convolutional neural network (CNN) on LeNet-5 over the MNIST dataset as a demo, and also compares the hardware resource of SMURF with the Taylor-series approximation and LUTs. Finally, in Section V, we conclude the paper and describe the prospects for future work.

\section{Preliminaries}

\subsection{Stochastic computing (SC)}
\begin{figure}[t]
\centering
\includegraphics[scale=0.3]{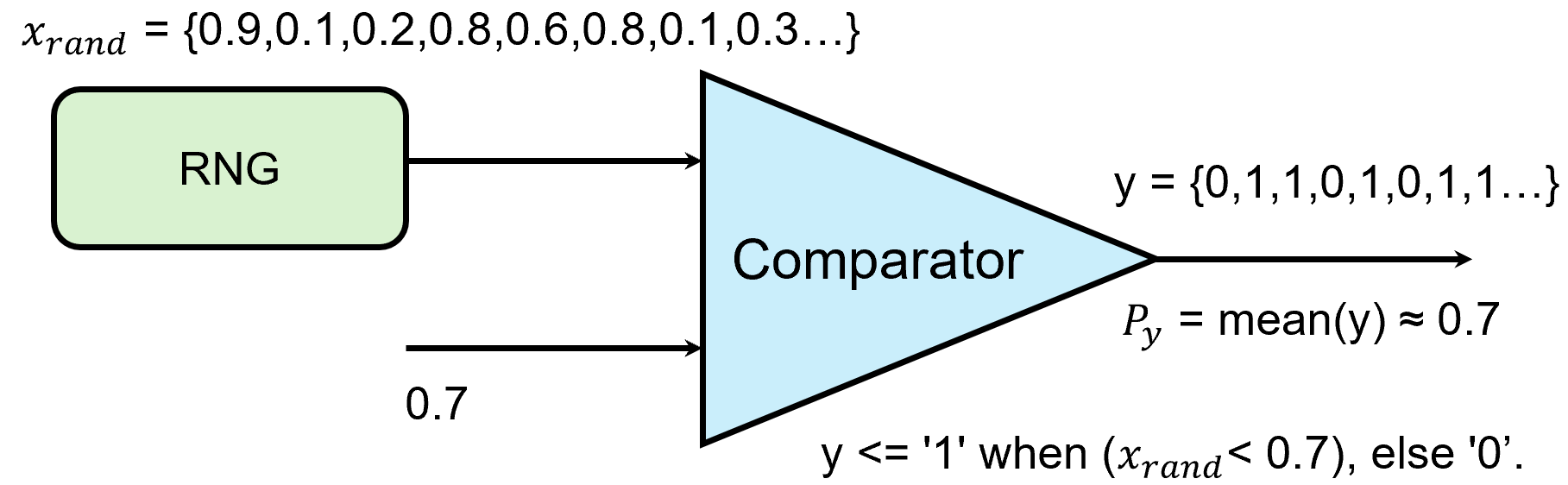}
\captionsetup{justification=centering}
\caption{The architecture of a stochastic number generator (SNG).}
\label{fig_SNG}
\end{figure}

SC is a computational approach based on random bitstream representation, which differs from conventional fixed-point binary computing. In conventional fixed-point arithmetic, numbers are represented using a fixed number of bits, with a specific number of bits reserved for the integer part and the remaining bits for the fractional part. In SC, each number or data is represented by a random value, typically generated via a stochastic number generator (SNG) that produces `0's and `1's. Statistical properties of random bit sequences can be leveraged to obtain computational results. 

As shown in Fig.~\ref{fig_SNG}, an SNG is utilized to convert a floating-point number into a stochastic number (SN). An SNG consists of a binary comparator and a random number generator (RNG). For example, we generate the stochastic number for 0.7 by setting the threshold of the comparator to this value. We generate one random number $x_{rand}$ ranged $[0,1]$ from the RNG and compare this number with the threshold. If $x_{rand}<0.7$, the SNG outputs a `1' or else a `0'. The mean of the generated bitstream approaches 0.7 when the length of the bitstream gets longer according to the law of large numbers. To convert (approximate) an SN back to its input value, we count the number of `1's in a bitstream via a binary counter and take the average. 
\begin{figure}[t]
\centering
\includegraphics[scale=0.3]{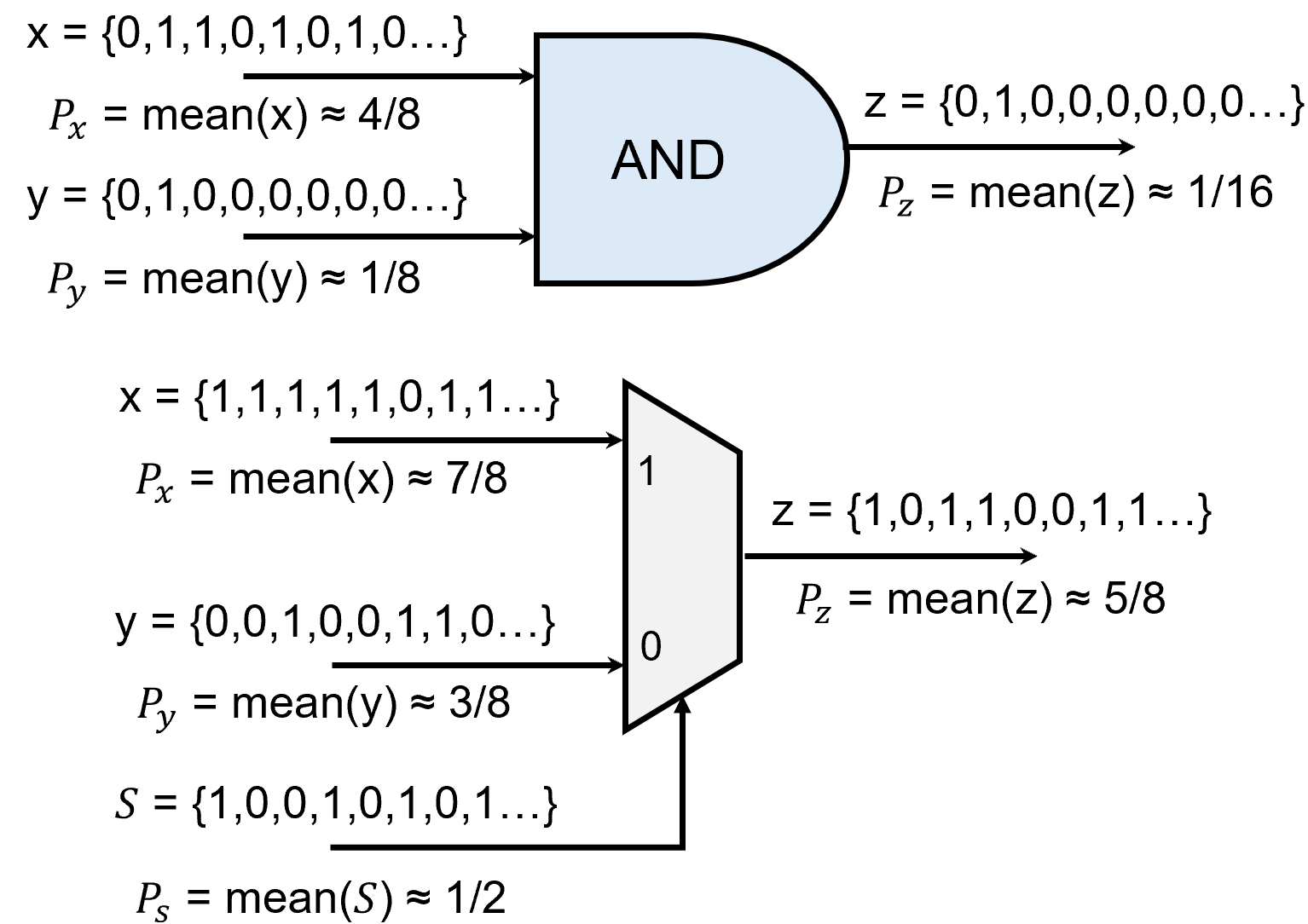}
\captionsetup{justification=centering}
\caption{Stochastic multiplication and addition.}
\label{fig_stochastic}
\end{figure}
Fig.~\ref{fig_stochastic} shows SC examples of binary multiplication and addition assuming all inputs and outputs are normalized to $[0,1]$. In the binary random bitstream format, we only need an AND gate to perform a stochastic multiplication since the product between two bitstreams follows the AND operation at their corresponding positions. Here, $x$ and $y$ are two independent random bitstreams representing two prescribed stochastic values, and $z$ is the output bitstream. Denoting the means of $x$, $y$ and $z$ as $P_x$, $P_y$ and $P_z$, respectively, we have $P_z=P_x P_y$. For stochastic addition, a scaling bitstream $s$ of probability $P_{s}=1/2$ and a MUX are in place. Using the notational convention of $\overline{P_s}=1-P_s$, it is readily seen that $P_z=P_s P_x+\overline{P_s}P_y=(P_x+P_y)/2$, which can then be multiplied by $2$ (using zero-cost hardware left-shifting) to restore the sum.

\begin{figure}[t]
\centering
\includegraphics[scale=0.3]{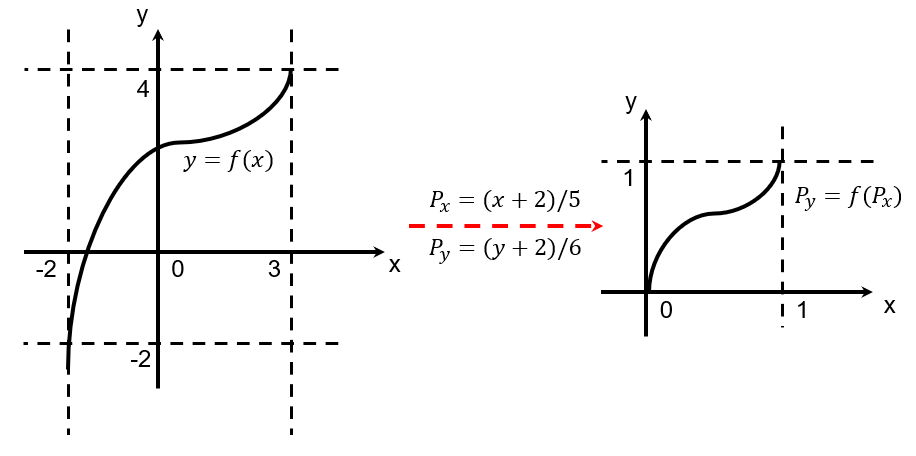}
\captionsetup{justification=centering}
\caption{Mapping the variables of a function to the spatial domain.}
\label{mapping}
\end{figure}
Although SC is typically used for computing SN values in $[0,1]$, we can apply SC to compute any function by normalizing its input-output variables to $[0,1]$. This normalization step can be done by a simple bijective linear transformation. An example is in Fig.~\ref{mapping} which illustrates the mapping of a function with input-output ranges $[-2,3]$ and $[-2,4]$. After the SC computation, we can map the output value back to its original range by linear scaling. Moreover, in a convolution process, one can utilize SC for point-wise matrix multiplications (PwMMs) and apply SC to activation functions~\cite{19}, resulting in significantly improved energy efficiency compared to fixed-point multiplications.

\subsection{Random sampling gates}
Random sampling gates~\cite{20} can be used to implement any function with known probability distributions. The logic gate used to map one probability distribution directly to another is called the $\theta$-gate. The logic gate used to map one probability distribution to another under a condition is called the conditional probability table (CPT) gate.

\textbf{$\theta$-gate~\cite{20}} samples a random number set according to a certain probability to obtain a binary Bernoulli distribution. By comparing the value between a prescribed input value and a random entropy source, one obtains a binary (${0,1}$) output value. The gate can sample exactly from each given distribution and produce one sample per cycle, e.g., the SNG in Fig.~\ref{fig_SNG} is a simple $\theta$-gate. As the sequence becomes sufficiently long, the output of a $\theta$-gate approaches the set distribution. This is also equivalent to the Monte Carlo method. A $\theta$-gate can also sample complex probability distributions such as the Sobol sequences. Henceforth, for convenience in further explanations, we refer to all gates that convert a full-precision value into an SN as $\theta$-gates.

\textbf{CPT-gate~\cite{20}} is a collection of $\theta$-gates, together with a MUX to select one of the $\theta$-gates as its output. This is equivalent to rolling a biased dice and choosing which dice value to take based on the input. In practice, we can adjust the distribution of the overall system output by modifying the threshold values of the $\theta$-gates. The control of the MUX often involves the use of random devices such as an FSM whose state transitions follow specific probabilistic rules.

\subsection{FSM-based nonlinear function generator}
\begin{figure}[t]
\centering
\includegraphics[scale=0.35]{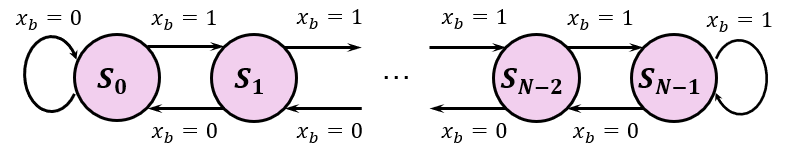}
\captionsetup{justification=centering}
\caption{The architecture of a chained $N$-state FSM where $x_b$ denotes the current bitstream (binary) value from a $\theta$-gate.}
\label{chain_fsm}
\end{figure}
In 2001, Brown and Card \cite{14} proposed that an $N$-state FSM can set the output of each state as either 0 or 1 to achieve a desired probability distribution when it reaches a steady state. Fig.~\ref{chain_fsm} shows an example FSM where the current state will transition to the right (or stay put at the rightmost) when the logical value $x_b$ is 1, and vice versa. In an $N$-state ($N$ even) FSM, if the outputs of the $1$st to $N/2$th states are 0 and those of the $(N/2)+1$th to $N$th states are 1, the overall output distribution of the FSM follows a $\tanh$ distribution. Let $x$ and $y$ be the SNs of the input and output variables respectively, the relationship between $P_x$ and $P_y$ is
\begin{equation}
    P_y\approx \frac{e^{\frac{N}{2}P_x}-e^{-\frac{N}{2}P_x}}{e^{\frac{N}{2}P_x}+e^{-\frac{N}{2}P_x}}
\end{equation}
\begin{figure*}[t]
  \centering
  \subfloat[]{
  \includegraphics[width=0.25\textwidth]{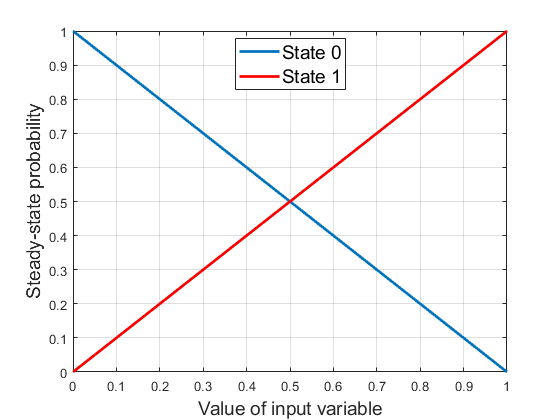}}
  \subfloat[]{
  \includegraphics[width=0.25\textwidth]{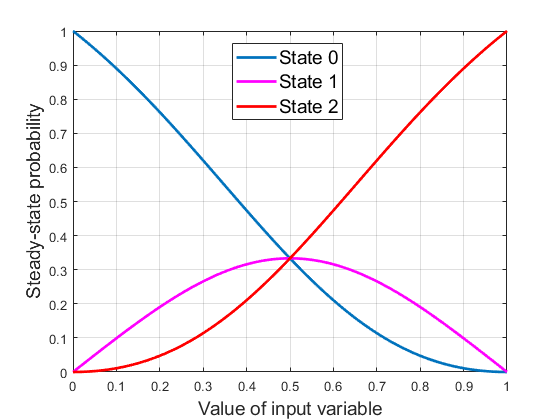}}
  \subfloat[]{
  \includegraphics[width=0.25\textwidth]{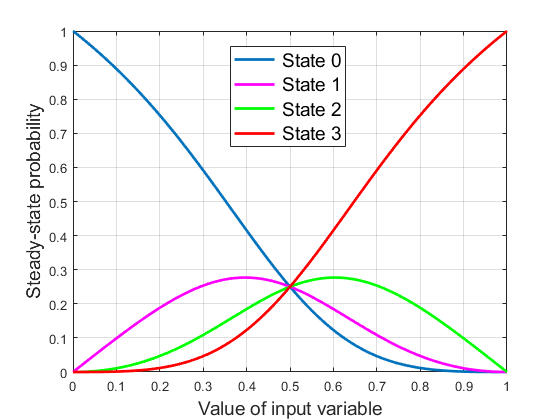}}
  \subfloat[]{
  \includegraphics[width=0.25\textwidth]{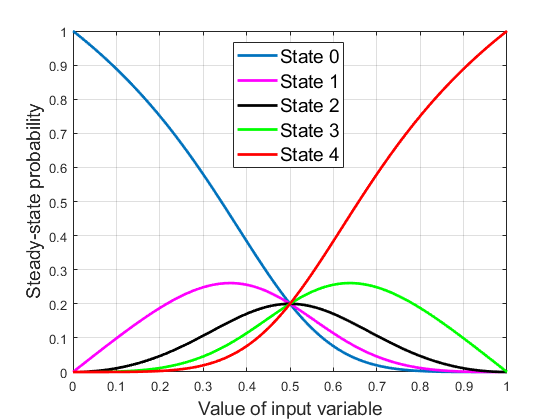}}
  \caption{(a)-(d) The steady-state probabilities of 2-, 3-, 4-and 5-state FSMs.}
  \label{steady_state}
\end{figure*}
This paper reveals that if the output of an FSM is not limited to simple probability distributions, but follows more complex distributions instead, we can leverage the FSM to compute various nonlinear functions other than just $\exp$ and $\tanh$. Due to the finite number of states in an $N$-state FSM used for nonlinear function approximation and the transiting probabilities from each state to another fall inside $[0,1]$, the state transition is traversable and aperiodic. In fact, the state transition process can be regarded as a steady-state Markov process, namely, one state has equal transition probabilities for transiting out to an adjacent state and transiting in from that state when the process reaches equilibrium~\cite{21}. Fig.~\ref{steady_state} depicts the steady-state probabilities of all states in 2-, 3-, 4- and 5-state FSMs.

The gradient difference between steady-state probabilities indicates the relative preference (i.e., visiting frequencies) of certain states in the long-term operation of the system. It can be seen from Fig.~\ref{steady_state} that when an FSM exhibits gradient differences between its states, it is capable of performing computations for input values in the range $[0,1]$. The states closer to the edges cover a wider range of output probability across $[0,1]$, while the states closer to the center correspond to a smaller range. Despite the smaller output range of the central states, they can still contribute to fine-tuning the overall output probability of the system, thereby increasing computing precision.

Referring to Fig.~\ref{steady_state}, when the input value increases, the steady-state probability of the leftmost state decreases from 1 to 0, while that of the rightmost state increases from 0 to 1. Those middle states exhibit a lower humps with different shifts. We remark that it is impossible to fit a nonlinear function with only two states due to their completely linear steady-state probabilities. Subsequently, a minimum of 3 states are required and increasing the number of states does not significantly improve the computation accuracy. In our examples, 4-state chains work well in all practical cases we have tested.

\section{SMURF Implementation}
This section presents the architecture and derivation of steady-state probabilities. The parameter calculation of a bivariate SMURF is showcased, whose generalization to the multivariate scenario is straightforward. 

\subsection{Architecture}
\begin{figure}[t]
\centering
\includegraphics[scale=0.5]{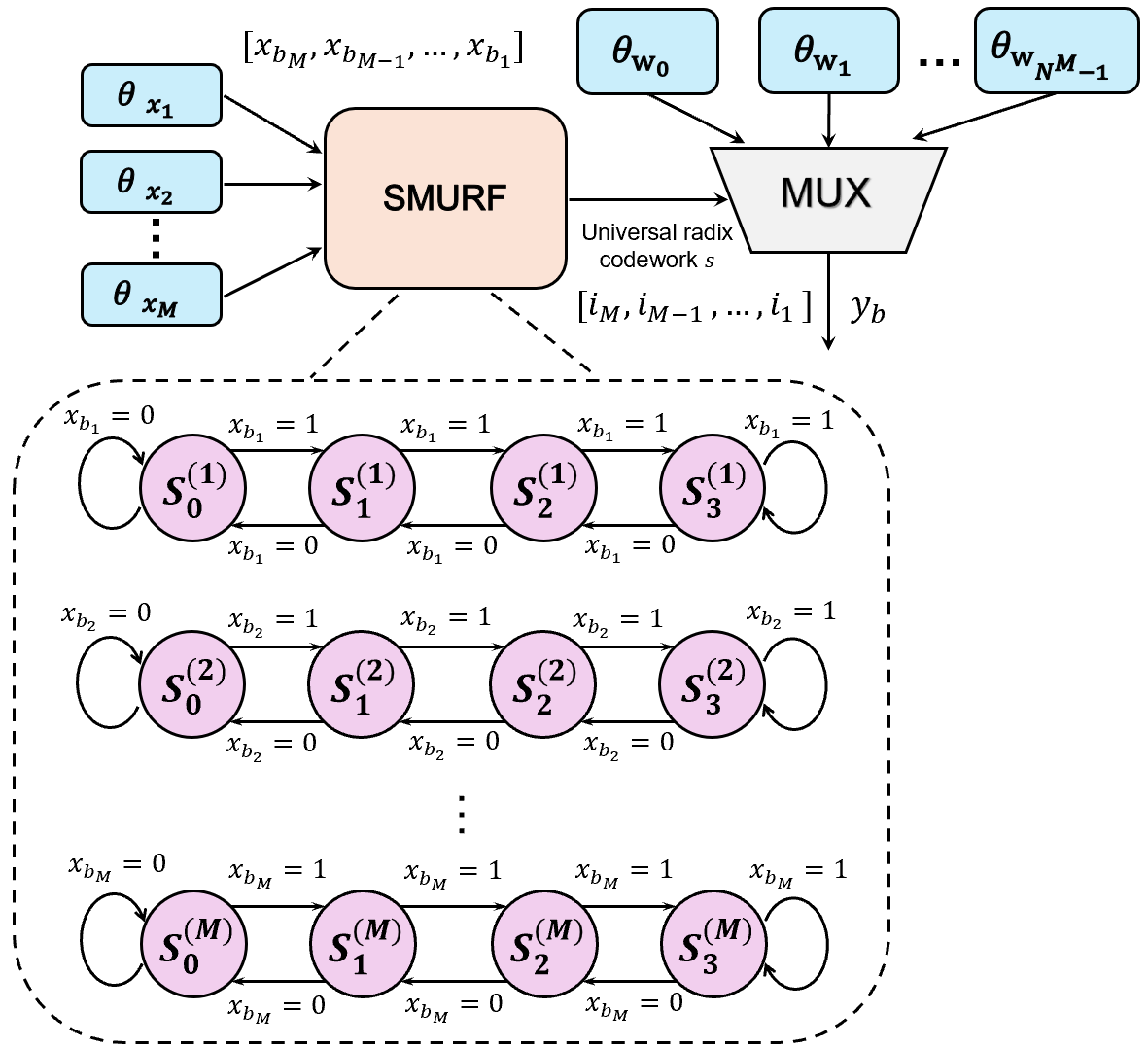}
\captionsetup{justification=centering}
\caption{Overarching architecture of a SMURF nonlinear function approximator.}
\label{fig:SMURF}
\end{figure}
SMURF states are synchronous to the clock signal and thus built on Moore FSMs. For ease of illustration, we set the number of states in the chained FSM affiliated with each of the $M$ input variables to be $N$, as shown in Fig.~\ref{fig:SMURF}. We then concatenate the states into one \textit{universal-radix} codeword $s=[S^{(M)}_{i_M},S^{(M-1)}_{i_{M-1}},\ldots ,S^{(1)}_{i_1}]$, or simply $s=[i_M, i_{M-1},\ldots ,i_1]$ wherein each `digit' spans $[0,N-1]$. This $s$ forms the select signal to the multiplexer (MUX), which is dubbed universal-radix since the radix varies with $N$, and can even be different for each FSM. When we use the same $N$ across different digits of the codeword $s$, it represents an `aggregate state' by itself with a total of $N^M$ (aggregate) states.

In practice, to realize the nonlinear generator in Fig.~\ref{fig:SMURF}, we instantiate a single RNG to output different random number sequences to be fed into the $\theta$-gates and the CPT-gate. Specifically, the random sequence from the RNG (cf. Fig.~\ref{fig_SNG}) is branched into differently delayed versions, emulating distinct pseudo-random sequences, that go into respective $\theta$-gates. The number of $\theta$-gates required in a CPT-gate corresponds to the number of states encoded by the codeword $s$, i.e., $N^M$.

In a clock cycle, the SMURF input consists of an $M$-bit binary codeword $[x_{b_M},x_{b_{M-1}},\ldots ,x_{b_1}]$ driving the $M$ ($N$-state) FSMs inside the SMURF to transit one state, thus producing the aforementioned codeword $s$. We assume the probabilities of input variables being 1 are $P_{x_1},P_{x_2},...,P_{x_M}$, respectively. The state transition of the $j$th FSM is then controlled by $P_{x_j}$ independently: when $x_{b_j}=1$, the state transits from $S^{(j)}_{i_j}$ to $S^{(j)}_{i_j+1}$, and from $S^{(j)}_{i_j}$ to $S^{(j)}_{i_j-1}$ when $x_{b_j}=0$. After each state transition, the updated codeword $s=[i_M, i_{M-1},\ldots ,i_1]$ controls the MUX of the CPT-gate to select the corresponding output bit of the chosen $\theta_{w_t}$-gate ($t=0,\cdots,N^M-1$). The arithmetic mean $P_y$ of the output binary bitstream $y_{b}$ then approximates the output value of our targeted nonlinear function.

\subsection{Bivariate SMURF}
\label{subsec:bivariate_SMURF}
We first walk through the design of a bivariate (i.e. $M=2$) SMURF given a bivariate nonlinear function. Let $i_{1}$, $i_{2}$ denote the state number of a current position in the first and second FSMs, i.e., $s=[i_2,i_1]$ and let $P_s=P_{[i_{2},i_{1}]}$ denote the joint steady-state probability of states $S^{(2)}_{i_{2}}$ and $S^{(1)}_{i_{1}}$. The probability of $S^{(1)}_{i_{1}}$ transiting to the right is then $P_{x_1}$, and that to the left is $\overline{P_{x_1}}=1-P_{x_1}$, etc. At steady-state equilibrium, we have:
\begin{subequations}
\begin{align}
P_{[i_{2},i_{1}+1]} \overline{P_{x_1}}&=P_{[i_{2},i_{1}]} P_{x_1},\\
P_{[i_{2}+1,i_{1}]} \overline{P_{x_2}}&=P_{[i_{2},i_{1}]} P_{x_2}.%
\end{align}%
\label{steady_21}%
\end{subequations}%
Defining the right-to-left-transit ratios $t_{x_1}=P_{x_1}/ \overline{P_{x_1}}$ and $t_{x_2}=P_{x_2}/\overline{P_{x_2}}$, (\ref{steady_21}) can be expressed as $P_{[i_{2},i_{1}+1]}= P_{[i_{2},i_{1}]} t_{x_1}$ and $P_{[i_{2}+1,i_{1}]}=P_{[i_{2},i_{1}]} t_{x_2}$. Noting that $P_{[0,0]}$ represents the output probability of the initial state, we then have $P_{[i_{2},i_{1}]}=P_{[0,0]}t^{i_2}_{x_2}t^{i_i}_{x_1}$. As the sum of all $P_{[i_{2},i_{1}]}$'s equals 1, we have:
\begin{equation}
   \sum^{N-1}_{i_{2}=0}\sum^{N-1}_{i_{1}=0}P_{[i_{2},i_{1}]}=P_{[0,0]} \sum^{N-1}_{i_2=0}t^{i_2}_{x_2}\sum^{N-1}_{i_1=0}t^{i_1}_{x_1}=1.
    \label{sum_1}
\end{equation}
Dividing both sides of the second equality in (\ref{sum_1}) with the summation terms to cast $P_{[0,0]}$ in terms of $t^{i_2}_{x_2}$ and $t^{i_1}_{x_1}$, we get%
\begin{equation}
    P_{[i_2,i_1]}=P_{[0,0]} t^{i_2}_{x_2}t^{i_1}_{x_1} =\frac{t^{i_2}_{x_2}t^{i_1}_{x_1}}{\sum^{N-1}_{i_2=0}t^{i_2}_{x_2}\sum^{N-1}_{i_1=0}t^{i_1}_{x_1}}.%
    \label{eqn:P_state}%
\end{equation}
Generally, when the state transition rules change, the specifics of (\ref{steady_21}) may vary, but (\ref{eqn:P_state}) remains unchanged. 

Increasing the bitstream length of $y_b$ (cf. Fig.~\ref{fig:SMURF}), the output value $P_y$ converges to a stable value. We can thus adopt the same convex optimization as in~\cite{18} to solve the parameters $w_0, \cdots, w_{N^2-1}$ with respect to a (bivariate) nonlinear function $f(x_1,x_2)$ which we approximate with a \textit{target function} $T(P_{x_1}, P_{x_2})$ using SMURF. Specifically, if the input SNs are $P_{x_1}$ and $P_{x_2}$ with their joint probability $P_{x_{1}x_{2}}$, this convex optimization problem is equivalent to minimizing%
\begin{equation}
    \epsilon=\int^{1}_{0}\int^{1}_{0}(T(P_{x_{1}},P_{x_{2}})-P_{y})^2d(P_{x_{1}})d(P_{x_{2}}).%
    \label{epsilon1}
\end{equation}
Expanding~(\ref{epsilon1}) yields
\begin{equation}
    \begin{aligned}
\epsilon&=\int^{1}_{0}\int^{1}_{0}T(P_{x_{1}},P_{x_{2}})^2d(P_{x_{1}})d(P_{x_{2}}) \\
&-2\int^{1}_{0}\int^{1}_{0}T(P_{x_{1}},P_{x_{2}}) P_{y} d(P_{x_{1}})d(P_{x_{2}}) \\
&+\int^{1}_{0}\int^{1}_{0}P_{y}^{2}d(P_{x_{1}})d(P_{x_{2}}).%
\label{epsilon2}
        \end{aligned}
\end{equation}
Since the first term of~(\ref{epsilon2}) is a constant, the goal is simplified to minimizing the last two terms only.
Setting
\begin{equation}
    \textbf{b}=[P_{w_{0}},P_{w_{1}},...,P_{w_{N^2-1}}]^T%
    \label{b}
\end{equation}
which is our variable column vector to be solved, and the row vector%
\begin{equation}
\begin{aligned}
    \textbf{c}&=[-\int^{1}_{0}\int^{1}_{0}T(P_{x_{1}},P_{x_{2}}) P_{[0,0]} d(P_{x_{1}})d(P_{x_{2}}),\\
    &-\int^{1}_{0}\int^{1}_{0}T(P_{x_{1}},P_{x_{2}}) P_{[0,1]} d(P_{x_{1}})d(P_{x_{2}}),...,\\
    &-\int^{1}_{0}\int^{1}_{0}T(P_{x_{1}},P_{x_{2}}) P_{[N-1,N-1]} d(P_{x_{1}})d(P_{x_{2}})]%
    \label{c}
    \end{aligned}
\end{equation}
which is deterministic since the double integration can be computed in the domain $[0,1]\times [0,1]$ with the prescribed $T(P_{x_1}, P_{x_2})$ and the closed-form expression of $P_{[i_2, i_1]}$ from (\ref{eqn:P_state}). Now further setting up the deterministic $N^2\times N^2$ matrix%
\begin{equation}
    \textbf{H}=[H_{[0,0]},H_{[0,1]},...,H_{[N-1,N-1]}],
    \label{H}
\end{equation}
wherein the stacked columns are
\begin{equation}
\begin{aligned}
    H_{s} &=[\int^{1}_{0}\int^{1}_{0}P_{s} P_{[0,0]}d(P_{x_{1}})d(P_{x_{2}}),\\
    &\int^{1}_{0}\int^{1}_{0}P_{s} P_{[0,1]}d(P_{x_{1}})d(P_{x_{2}}),...,\\
    &\int^{1}_{0}\int^{1}_{0}P_{s} P_{[N-1,N-1]}d(P_{x_{1}})d(P_{x_{2}})]^T,
    \label{Ht}
    \end{aligned}
\end{equation}
where the codeword $s$ takes on $[0,0], [0,1], \ldots, [N-1, N-1]$, and $\textbf{H}$ is by construction symmetric. Thus, the convex optimization is transformed to finding the minimum of 
\begin{equation}
\varphi=\textbf{b}^{T}\textbf{H}\textbf{b}+2\textbf{c}\textbf{b}.
    \label{fai_convex}
\end{equation}
Such optimization solves the vector $\textbf{b}$ and subsequently its entries $P_{w_{t}}$'s for setting the $w_{t}$'s in the corresponding $\theta$-gate inputs going into the MUX. 

\subsubsection{Example 1} We examine a 2D Euclidean distance function $f(x_1,x_2)=\sqrt{x_1^2+x_2^2}$. In this case, we treat the variables $x_1$ and $x_2$ as SNs $P_{x_1}$ and $P_{x_2}$, respectively, with the target function%
\begin{align}
T(P_{x_1},P_{x_2})=\sqrt{P_{x_1}^2+P_{x_2}^2},~0\leq P_{x_1}, P_{x_2}\leq 1.
    \label{Euclidean_stochastic}
\end{align}
Setting $N=4$, we use (\ref{fai_convex}) to compute the corresponding $N^2=16$ $w_{t}$'s for the target function, listed in Table~\ref{para_Euclidean}.
\begin{table}[t]
\caption{$w_{t}$'s for computing $\sqrt{x_1^2+x_2^2}$.}
\centering
 \setlength{\tabcolsep}{1.5mm}
\begin{tabular}{cccc}
\toprule
\multicolumn{1}{c|}{$P_{w_0}=0$}    & \multicolumn{1}{c|}{$P_{w_1}=0.6083$}    & \multicolumn{1}{c|}{$P_{w_2}=0.0474$}    & $P_{w_3}=0.6911$    \\ \midrule
\multicolumn{1}{c|}{$P_{w_4}=0.6083$}    & \multicolumn{1}{c|}{$P_{w_5}=0.3749$} & \multicolumn{1}{c|}{$P_{w_6}=0.4527$} & $P_{w_7}=0.8372$    \\ \midrule
\multicolumn{1}{c|}{$P_{w_8}=0.0474$}    & \multicolumn{1}{c|}{$P_{w_9}=0.4527$} & \multicolumn{1}{c|}{$P_{w_{10}}=0.0159$} & $P_{w_{11}}=0.5946$ \\ \midrule
\multicolumn{1}{c|}{$P_{w_{12}}=0.6911$} & \multicolumn{1}{c|}{$P_{w_{13}}=0.8372$}    & \multicolumn{1}{c|}{$P_{w_{14}}=0.5946$}    & $P_{w_{15}}=0.9846$    \\ \bottomrule                                                            
\end{tabular}
\label{para_Euclidean}
\end{table}

\subsubsection{Example 2}  Ref~\cite{22} applied convolution using the Hartley stochastic computing (HSC) in the convolutional layers of a CNN, converting the filters and input feature maps from the spatial domain to the frequency domain via Hartley Transform (HT). The HT for a $Q\times Q$ matrix is expressed as 
\begin{equation}
\begin{aligned}
    H_{2\pi}(k,l)=\frac{1}{Q}\sum^{Q-1}_{m=0}\sum^{Q-1}_{n=0}f[m,n]
    \cas\left(\frac{2\pi}{Q}(km+ln)\right),
\end{aligned}
    \label{DFT}
\end{equation}
where $\cas(x)=\sin(x)+\cos(x)$, $m$ and $n$ are the row and column of the corresponding element, $k$ and $l$ are two coefficients. We represent $f[m,n]$ in (\ref{DFT}) with $\sin(x_1)$, and consider $x_2=\frac{2\pi}{Q}(km+ln)$, thereby turning~(\ref{DFT}) into a simplified form: 
\begin{equation}
\begin{aligned}
H_{2\pi}(k,l)&=\sin(x_1) \cas(x_2).
\label{sin}
\end{aligned}
\end{equation}
Then, we convert the variables to SNs $P_{x_1}$ and $P_{x_2}$ such that the target function becomes%
\begin{equation}
    \begin{aligned}
        T(P_{x_1},P_{x_2}) = \sin(P_{x_1})\cas(P_{x_2}),
        ~0\leq P_{x_1}, P_{x_2}\leq 1.
        \label{HT_stochastic}
    \end{aligned}
\end{equation}
The parameters $w_t$'s used to implement the HT function are listed in Table~\ref{para_HT}. 

\begin{table}[t]
\caption{$w_{t}$'s for computing  $\sin(x_1)\cas(x_2)$.}
\centering
\setlength{\tabcolsep}{1.5mm}
\begin{tabular}{cccc}
\toprule
\multicolumn{1}{c|}{$P_{w_0}=0$}    & \multicolumn{1}{c|}{$P_{w_1}=0.4002$}    & \multicolumn{1}{c|}{$P_{w_2}=0.4002$}    & $P_{w_3}=0.3379$    \\ \midrule
\multicolumn{1}{c|}{$P_{w_4}=0.3379$}    & \multicolumn{1}{c|}{$P_{w_5}=0.4334$} & \multicolumn{1}{c|}{$P_{w_6}=0.4334$} & $P_{w_7}=0.6600$    \\ \midrule
\multicolumn{1}{c|}{$P_{w_8}=0$}    & \multicolumn{1}{c|}{$P_{w_9}=0.5407$} & \multicolumn{1}{c|}{$P_{w_{10}}=0.5407$} & $P_{w_{11}}=0.4564$ \\ \midrule
\multicolumn{1}{c|}{$P_{w_{12}}=0.4564$} & \multicolumn{1}{c|}{$P_{w_{13}}=0.5854$}    & \multicolumn{1}{c|}{$P_{w_{14}}=0.5854$}    & $P_{w_{15}}=0.8916$    \\ \bottomrule    
\end{tabular}
\label{para_HT}
\end{table}

\subsection{Multivariate SMURF}
\label{subsec:multivariate_SMURF}
Here we extend SMURF to its general multivariate setting. When the number of input variables is $M$, we denote the FSM number as $m$ with its transit ratio (cf.~Section~\ref{subsec:bivariate_SMURF}) 
\begin{align}
t_{x_m}=P_{x_m}/\overline{P_{x_m}},~m=1,2,\ldots,M. 
\end{align}
We have
\begin{align}
P_{[i_M,\ldots,i_m+1,\ldots,i_1]}=P_{[i_M,\ldots,i_m,\ldots,i_1]} t_{x_m},
\end{align}
Assume without loss of generality that all FSMs have $N$ states, since the sum of probabilities of all states equals 1, we get%
\begin{align}
    \sum^{N-1}_{i_M=0}\sum^{N-1}_{i_{M-1}=0}\ldots\sum^{N-1}_{i_1=0}P_{[i_M,i_{M-1},\ldots,i_1]}=1,
\end{align}
Expanding all terms yields
\begin{align}
    P_{[0,0,...,0]}\sum^{N-1}_{i_{M}=0}t^{i_M}_{x_M}\sum^{N-1}_{i_{M-1}=0}t^{i_{M-1}}_{x_{M-1}}...\sum^{N-1}_{i_{1}=0}t^{i_1}_{x_1}=1,%
    \label{eqn:P_init}
\end{align}
where $P_{[0,0,...,0]}$ represents the output probability of the initial state.
For the state $s=[i_M,i_{M-1},\ldots,i_1]$, we have%
\begin{align}
    P_{s}=P_{[0,0,...,0]} t^{i_M}_{x_M} t^{i_{M-1}}_{x_{M-1}}\ldots t^{i_1}_{x_1}.
\end{align}
Re-expressing $P_{[0,0,...,0]}$ with (\ref{eqn:P_init}),
\begin{equation}
    P_{s}=\frac{t^{i_M}_{x_M} t^{i_{M-1}}_{x_{M-1}}\ldots t^{i_1}_{x_1}}{\sum^{N-1}_{i_M=0}t^{i_M}_{x_M} \sum^{N-1}_{i_{M-1}=0}t^{i_{M-1}}_{x_{M-1}} \ldots\sum^{N-1}_{i_1=0}t^{i_1}_{x_1}}.
\end{equation}
The rest follows from the bivariate SMURF of solving~(\ref{fai_convex}), but with the size of the unknown vector $\textbf{b}$ now being $N^M$ and the order of integration in~(\ref{epsilon2}) now being $M$.

\subsubsection{Example 1} The softmax function is widely used for multi-class classification. This function allows us to transform raw data scores into a probability distribution. We examine the softmax of 3 inputs. By expressing the input variables as SNs $0\leq P_{x_1},P_{x_2},P_{x_3}\leq1$, the target function reads
\begin{align}
T(P_{x_1},P_{x_2},P_{x_3})=\frac{[\exp({P_{x_1}}),\exp({P_{x_2})},\exp({P_{x_3}})]}{\sum_{a=1}^{3}\exp(P_{x_{a}})}.%
\end{align}
The rest of the flow resembles that of Section~\ref{subsec:bivariate_SMURF}, but we omit the 64-entry table due to space constraint.

\subsubsection{Discussion} From the above examples, SMURF can be interpreted as a simplistic SC-based neural network for approximating a ground-truth nonlinear function. When the structure of SMURF is sufficiently complex and the number of states is large enough, it can perform more sophisticated computations. This constitutes our ongoing research tasks.
\begin{table*}[t]
\centering
\caption{Operational Comparison of SMURF and CORDIC.}
\begin{tabular}{cccc}
\toprule
 & $\sqrt{x_1^2+x_2^2}$ & $\sin{x_1}\cas{x_2}$ & $\frac{\exp{x_1}}{\exp{x_1}+\exp{x_2}}$\\ \midrule
  CORDIC  & $2\times(\circ)^2+1\times\sqrt{(\circ)}$ & \makecell{$2\times\sin{(\circ)}+1\times\cos{(\circ)}$\\ $+1\times \mbox{Add}+1\times\mbox{Multiply}$} & \makecell{$2\times\exp{(\circ)}$\\ $+1\times \mbox{Add}+1\times\mbox{Divide}$} \\ \midrule
SMURF (\textbf{ours})  & $1\times\sqrt{(\circ)^2+(\circ)^2}$      & $1\times\left(\sin{(\circ)}\cas{(\circ)}\right)$      & $1\times\left(\frac{\exp(\circ)}{\exp{(\circ)}+\exp{(\circ)}}\right)$                                \\ \bottomrule
\end{tabular}
\label{CORDIC}
\end{table*}
Here we further contrast SMURF against CORDIC~\cite{11} which is commonly used for realizing univariate nonlinear functions such as $\exp$, $\sin$, hyperbolic functions, and power functions. To compute a multivariate function with CORDIC, the only approach is to realize the parts of the multivariate function involving different independent variables separately, and then combine them with standard arithmetic. Table~\ref{CORDIC} compares the operations of SMURF and CORDIC for several multivariate functions. In short, SMURF can realize multivariate nonlinear functions using \emph{one} generalized architecture, while CORDIC requires multiple computations of different univariate functions, rendering a significantly higher complexity in terms of both software and hardware. Consequently, SMURF directly addresses the poor generalizability of CORDIC to the multivariate scenario.

\section{Experimental results}
We conduct software and hardware evaluations of SMURF against competing schemes. We also demonstrate the use of SMURF in a convolutional neural network (CNN) classification task, followed by further hardware analyses.

\subsection{Performance of SMURF in generic nonlinearities}
Standard fixed-point representation is employed for $\theta$-gate inputs, whose quantization error is negligible compared to other sources of errors, such as the Markov process uncertainties and random bitstream averaging.

\begin{figure}[t]
\centering
\includegraphics[scale=0.09]{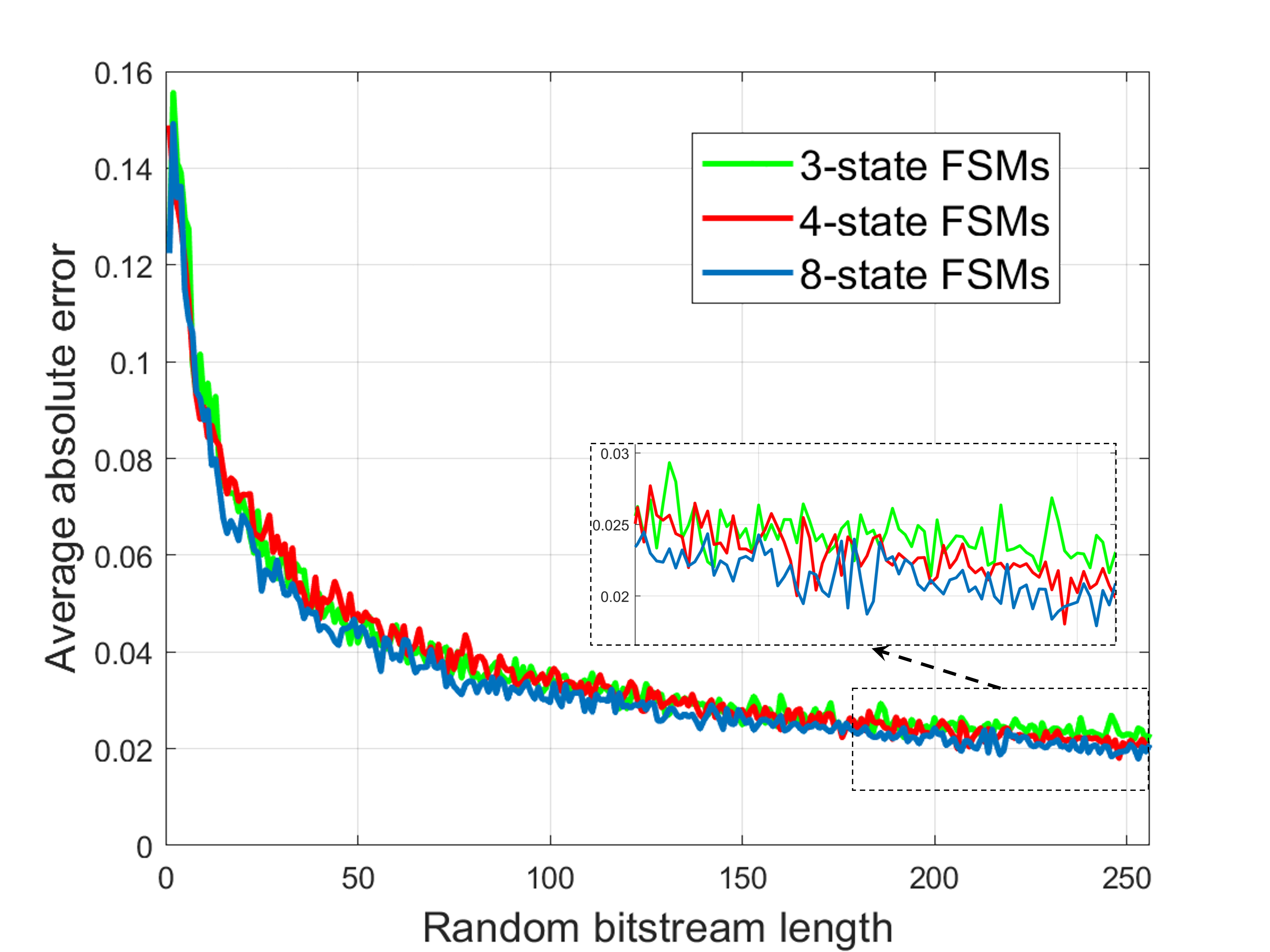}
\caption{SMURF approximation errors in a 3-variate softmax function with 3/4/8-state FSMs w.r.t. bitstream lengths.}
\label{256bits_vary}
\end{figure}
We first test the computing accuracy of SMURF in a 3-variate softmax function using 3-state, 4-state and 8-state FSMs for each input variable. The average absolute error with respect to the bitstream length is shown in Fig.~\ref{256bits_vary}, showing the fast decay and convergence of the average absolute errors. The average absolute errors are all around 0.15 near zero, and decrease to 0.02 at a bitstream length of 256. On the other hand, increasing the number of states in an FSM results in only small gains in output precision ($\leq$0.01). The main causes of the error are the stochastic process (dynamic stability) of the FSMs, and the instability introduced by the $\theta$-gates and CPT-gates (namely, fluctuations in mean computation), both arising from the finite-length random bitstreams. That said, in practical applications we often do not require a very high precision, e.g., prior work has shown CNNs can tolerate the relatively low accuracy of SC~\cite{19}. Balancing hardware complexity and accuracy, we fix the bitstream length at 64, whose average absolute error is around 0.04 in this example.
\begin{figure}[t]
\centering
\includegraphics[scale=0.3]{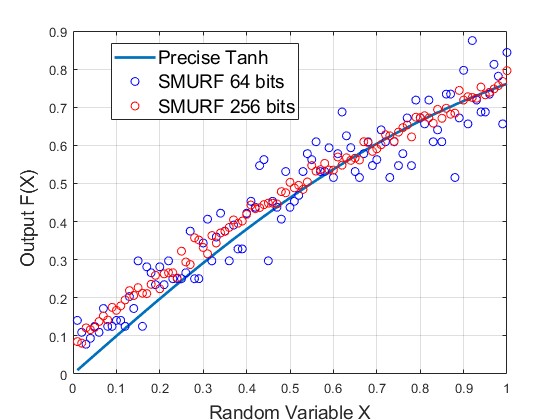}
\caption{SMURF approximation of $\tanh$.}
\label{tanh_scatter}
\end{figure}
\begin{figure}[t]
\centering
\includegraphics[scale=0.3]{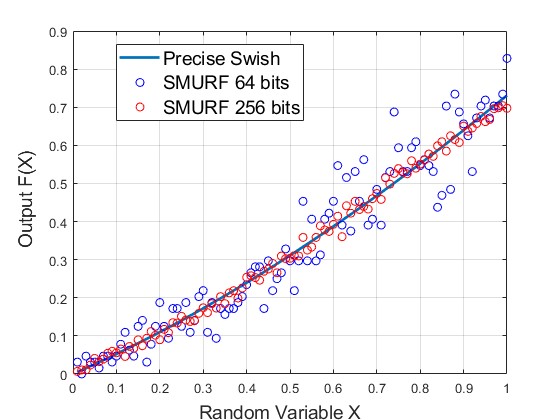}
\caption{SMURF approximation of $\swish$.}
\label{swish_scatter}
\end{figure}
Next, we approximate the common neural network activation functions, $\tanh$ and $\swish$, using SMURF with bitstream lengths of 64 and 256. Figs.~\ref{tanh_scatter} \&~\ref{swish_scatter} depict the approximation results. It can be seen that SMURF achieves a good approximation in both functions at the bitstream length of 64, and can highly restore these two nonlinearities at a length of 256. The average absolute errors approximating $\tanh$ ($\swish$) are 0.037 (0.033) and 0.011 (0.010) at 64 bits and 256 bits, respectively. In fact, the error at 64 bits is adequately small for neural network inferences.
\begin{figure*}[t]
  \centering
  \subfloat[][Euclidean distance]{
  \includegraphics[width=0.3\textwidth]{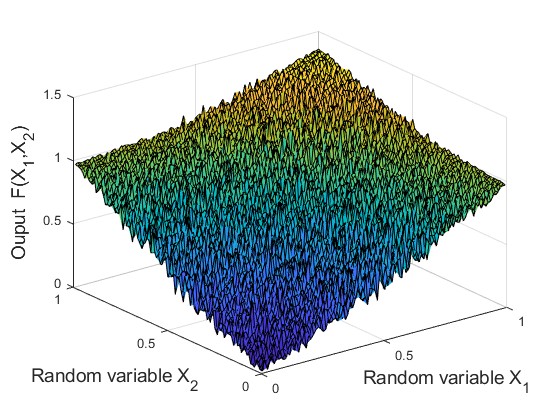}\label{fig_errorround}}
  \subfloat[][Hartley transform]{
  \includegraphics[width=0.3\textwidth]{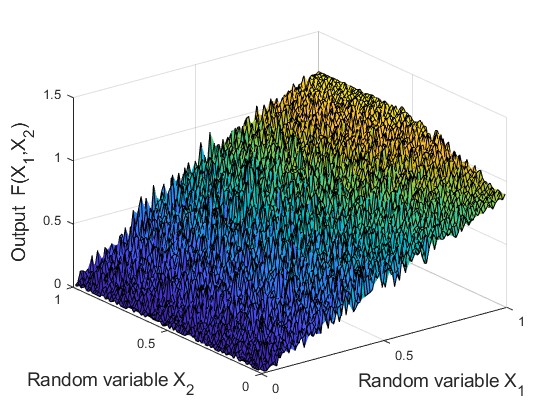}\label{fig_errorHT}}
  \subfloat[][Softmax (bivariate)]{\includegraphics[width=0.3\textwidth]{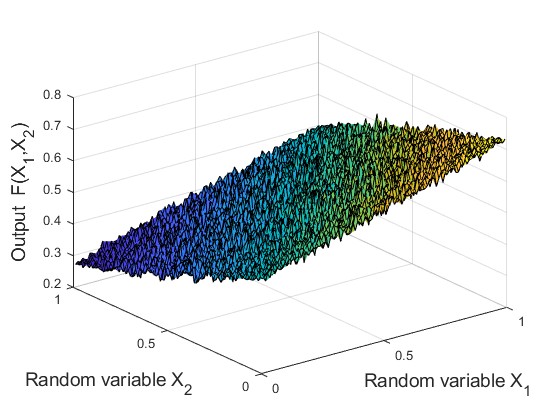}\label{fig_errorsoftmax}}
  \caption{The performance of SMURF approximating bivariate functions with the bitstream length of 64.}
  \label{bivariate_distribution}
\end{figure*}

Fig.~\ref{bivariate_distribution} shows the performance of SMURF approximating the Euclidean distance, HT and bivariate softmax with an input bitstream length of 64. In Figs.~\ref{fig_errorround} \&~\ref{fig_errorHT}, the average absolute errors are both around 0.032, whereas in Fig.~\ref{fig_errorsoftmax} the error is 0.014 which is smaller possibly due to the smoother nonlinearity. These results validate that SMURF exhibits generally good approximation accuracy in multivariate nonlinear functions without excessively long bitstreams.

\subsection{Performance of SMURF in a CNN}
To verify the use of SMURF in realizing the nonlinear activations in a CNN, we train a LeNet-5 for MNIST classification using SMURF approximators for the activation and HT functions, forming a new SC-based CNN. During training, we represent the weights, HT inputs and inputs to the activation using 64-bit random bitstreams. We then contrast the output accuracy of CNN/SMURF against CNN/HSC~\cite{22}. The fixed-point precision and the angular precision in the latter are 8 bits and 11 bits, respectively, plus a 128-bit bitstream in SC-PwMMs. For a fair comparison, we also set the bitstream length of SC-PwMMs in CNN/SMURF to 128.

\begin{table}[t]
\caption{MNIST test accuracies of vanilla CNN, CNN/HSC and CNN/SMURF.}
\centering
\setlength{\tabcolsep}{3.5mm}
\begin{tabular}{cccccc}
\toprule
                   & Vanilla CNN & CNN/HSC & CNN/SMURF \\ \midrule
Accuracy/\%              & 99.67           & 98.04                       & 98.42
\\ \bottomrule
\end{tabular}
\label{CNN_performance}
\end{table}
\begin{table}[t]
\centering
\caption{Hardware implementations of CNN/SMURF, CNN/HSC, and vanilla CNN.}
\resizebox{1.0\columnwidth}{!}{
\begin{tabular}{ccc}
\toprule
& Convolution             & Activation functions\\ \midrule
Vanilla CNN & standard convolution & standard $\ReLU$, softmax  \\ \midrule
CNN/HSC         & LUT-HT, SC-PwMM    & standard $\ReLU$, softmax  \\ \midrule 
CNN/SMURF       & SMURF-HT, SC-PwMM          & SMURF\\ \bottomrule
\end{tabular}}
\label{CNN_hardware}
\end{table}

Table~\ref{CNN_performance} shows the MNIST test accuracies of a vanilla and two SC-based LeNet-5, validating that SC-based operators are capable of running CNNs, with only a slight 1\% accuracy drop compared to the full-precision CNN in this example. Table~\ref{CNN_hardware} enumerates the implementation differences of these CNN schemes. In particular, CNN/HSC utilizes LUT-HTs and SC-PwMMs to perform convolution, thereby obviating the high-complexity full-precision adders and multipliers. We remark that CNN/HSC mainly implements PwMM using SC, and requires LUTs for HT. It is not mentioned in~\cite{22} how the nonlinear activations (e.g., ReLU, softmax, etc.) are done, but if they are not to be achieved via SC, then other approximators such as Taylor series or LUTs are further required. Differently, CNN/SMURF reduces hardware complexity by employing all SMURF-based SC for HT and also nonlinear activations. To get a firmer grasp of hardware complexities, we thereby compare the hardware costs of SMURF, LUT, and Taylor approximation in the next section.

\subsection{Hardware implementation}
To further validate the efficacy of SMURF, we benchmark the hardware metrics of SMURF deployment on the field-programmable gate array (FPGA) platform. The process node is Semiconductor Manufacturing International Corporation (SMIC) $65$nm, and the operating frequency of the system is 400MHz. We verify the correctness of the register transfer level (RTL) code and the network table through Modelsim. For fairness, the average absolute errors of all methods are equated to around 0.015. For an illustrative purpose, we implement the bivariate Euclidean distance function, which is expanded to a cubic Taylor-series polynomial on hardware in order to match the SMURF-HT accuracy at a bitstream length of 256. As before, we use two 4-state FSMs in SMURF-HT.

For power consumption, SMURF consumes 0.51mW, mostly due to the RNG. The total area of the module is 5294.72um$^2$. The RNG occupies approximately 1600um$^2$, and the SMURF core and CPT-gate occupy 104.4um$^2$ and 293.4um$^2$, respectively. The Taylor approximation error is determined by its expansion order and data bitwidth. In this experiment, we set the bitwidth to 16, and design a 4-stage pipeline. As shown in Table~\ref{hardware_nonlinear}, the power consumption of the Taylor scheme is 3.53mW. Due to the adders and multipliers, the hardware area consumption is significantly greater, which is 32941.44um$^2$. Moreover, experimental results show that the timing can barely reach 400MHz in the case of a 4-level pipeline, which has little potential to work at higher frequencies. Additionally, we conducted hardware simulation of LUTs using the same output bitwidth. As seen from Table~\ref{hardware_nonlinear}, although the power consumption of LUT is only 0.10mW, it incurs a significant area overhead of 238176.38um$^2$. 

\begin{table}[t]
\centering
\caption{Hardware metrics of SMURF, Taylor and LUT.}
\resizebox{1.0\columnwidth}{!}{
\begin{tabular}{cccc}
\toprule
Methods                 & Area/um$^2$ & Power/mW & Area$\cdot$ Power/um$^2\cdot$ mW \\ \midrule
SMURF               & 5294.72     & 0.51       & 2700.31              \\ \midrule
Taylor & 32941.44    & 3.53       & 116283.28            \\ \midrule
LUT                     & 238176.38   & 0.10       & 23817.64               \\ \bottomrule
\end{tabular}}
\label{hardware_nonlinear}
\end{table}
We highlight that the area of SMURF only accounts for 16.07\% of the Taylor approximation, while the power consumption of SMURF is 14.45\% of that using the Taylor approximation. Although the power consumption of SMURF is 5 times that of LUT, the area of SMURF is only 2.22\% of the latter. To comprehensively evaluate the area and power consumption of these three methods, we multiply these two metrics. The area$\cdot$power of SMURF is only 2.32\% and 11.34\% of that using Taylor approximation and LUT, respectively. From the results, we conclude that SMURF is superior to the conventional Taylor approximation in both energy efficiency and area efficiency. Although a LUT has a short critical path and lower power consumption, its area overhead is 45$\times$ that of SMURF. Therefore, overall, SMURF demonstrates superb performance in terms of hardware.

\subsection{Additional Remarks}
The proposed SMURF nonlinear function computing architecture exhibits good accuracy in computing various univariate and multivariate nonlinear functions. In the context of hardware implementation, SMURF significantly reduces hardware and power consumption versus the Taylor-series approximation and LUTs. Importantly, SMURF can efficiently compute various univariate and multivariate nonlinear functions using only one general architecture by simply changing the parameters, namely, in the $\theta$-gates. We believe that in scenarios requiring various nonlinear functions, SMURF will demonstrate superior adaptability. In other words, SMURF exhibits great versatility and offers convenience and energy efficiency in practical applications. 

\section{Conclusion}
This paper has proposed a novel stochastic computing architecture leveraging a \textbf{S}tochastic \textbf{M}ultivariate \textbf{U}niversal-\textbf{R}adix \textbf{F}inite-state machine, called SMURF, for generic nonlinear function approximation. Both its hardware construction and analytical derivation of parameters are detailed, backed by extensive experiments to showcase its superiority over Taylor-series approximation and lookup-table approaches. Experiments have shown that SMURF requires only 16.07\% area and 14.45\% power consumption of Taylor-series approximation, and only 2.22\% area of lookup-table schemes, making it highly competitive for edge AI. In the future, it would be highly desired to extend SMURF to more sophisticated finite-state machines with more complex behaviors, as well as to intrinsically handle multi-output nonlinear functions for wider application.
\ifCLASSOPTIONcaptionsoff
  \newpage
\fi








\end{document}